\documentclass[runningheads]{llncs}

\usepackage[T1]{fontenc}

\usepackage{amsmath}
\usepackage{amssymb}

\usepackage{graphicx}
\usepackage{xcolor}
\usepackage{booktabs}
\usepackage{multirow}
\usepackage{makecell}
\usepackage{tabularx}
\usepackage{array}
\usepackage{pifont}
\usepackage{subcaption}

\usepackage{algorithm}
\usepackage{algpseudocode}

\usepackage{hyperref}
\definecolor{myhexcolor0}{HTML}{FF00FF}
\definecolor{myhexcolor1}{HTML}{C13EFF}
\definecolor{myhexcolor2}{HTML}{8679FF}
\definecolor{myhexcolor3}{HTML}{42BDFF}

\begin{document}

\title{\textit{OsteoFlow}: Lyapunov-Guided Flow Distillation for Predicting Bone Remodeling after Mandibular Reconstruction}

\titlerunning{OsteoFlow: Lyapunov-Guided Flow Distillation}

\author{
Hamidreza Aftabi\inst{1,4}\textsuperscript{*}
\and
Faye Yu\inst{2}
\and
Brooke Switzer\inst{3}
\and
Zachary Fishman\inst{4}
\and
Eitan Prisman\inst{3}
\and
Antony Hodgson\inst{2}
\and
Cari Whyne\inst{4}
\and
Sidney Fels\inst{1}\textsuperscript{\ensuremath{\dagger}}
\and
Michael Hardisty\inst{4}\textsuperscript{\ensuremath{\dagger}}
}

\authorrunning{H. Aftabi et al.}

\institute{
Department of Electrical and Computer Engineering,
University of British Columbia,
Vancouver, British Columbia, Canada
\and
Department of Mechanical Engineering,
University of British Columbia,
Vancouver, British Columbia, Canada
\and
Department of Surgery,
University of British Columbia,
Vancouver, British Columbia, Canada
\and
Sunnybrook Research Institute,
University of Toronto,
Toronto, Ontario, Canada
}

\maketitle

\begin{center}
\small
\textsuperscript{*}Corresponding author:
\href{mailto:aftabi@student.ubc.ca}{aftabi@student.ubc.ca}.
\qquad
\textsuperscript{\ensuremath{\dagger}}Joint senior authors.
\end{center}

\begin{abstract}
Predicting long-term bone remodeling after mandibular reconstruction would be of great clinical benefit, yet standard generative models struggle to maintain trajectory-level consistency and anatomical fidelity over long horizons, particularly in low-data regimes. We introduce \textit{OsteoFlow}, a flow-based framework for predicting Year-1 postoperative CT scans from Day-5 scans. Our core contribution is Lyapunov-guided trajectory distillation. Unlike one-step distillation, our method distills a continuous trajectory over transport time from a registration-derived stationary velocity field teacher. Combined with a resection-aware image loss, this approach enforces geometric correspondence without sacrificing generative capacity. Evaluated on 344 paired regions of interest, \textit{OsteoFlow} significantly outperforms state-of-the-art baselines, reducing mean absolute error in the surgical resection zone by approximately \(20\%\). These results highlight the promise of trajectory distillation for long-term prediction in low-data clinical settings. Code is available at \url{https://github.com/hamidreza-aftabi/OsteoFlow}.

\keywords{Mandibular Reconstruction \and Flow-Based Modeling \and Surgical Planning \and Lyapunov-Guided Distillation}
\end{abstract}

\section{Introduction}
Mandibular reconstruction after segmental resection aims to restore jaw form and function in patients with oral and maxillofacial disease~\cite{aftabi2024computational}. Despite advances in surgical planning and fixation, bone healing at the graft-host interface remains a major clinical challenge. Nonunion at the graft-host interface occurs up to 37\% of cases in some sites~\cite{swendseid2020natural}, and can lead to pain, impaired function, hardware complications, and revision surgery. Although union is influenced by multiple patient and treatment-related factors (e.g., radiation exposure), previous studies suggest that local graft-host apposition (interface gap and contact geometry) is a key determinant of healing~\cite{aftabi2025osteoopt,sabiq2024evaluating}. This motivates local image-based prediction at the graft-host interface, where remodeling most directly relates to union.

Prior studies on mandible reconstruction prediction have largely used physics-based and biomechanical simulations, including bone remodeling~\cite{aftabi2024extent,aftabi2025optimizing,aftabi2026patient}. While mechanistically grounded and interpretable, these methods often require case-specific assumptions and parameterization, and can be computationally expensive at deployment. Even physics-informed machine learning models rely on extensive simulation guidance for bone remodeling, given the lack of a canonical governing PDE~\cite{wu2021machine}.
\begin{figure}[t] 
\centering \includegraphics[width=1\linewidth,height=.221\textheight]{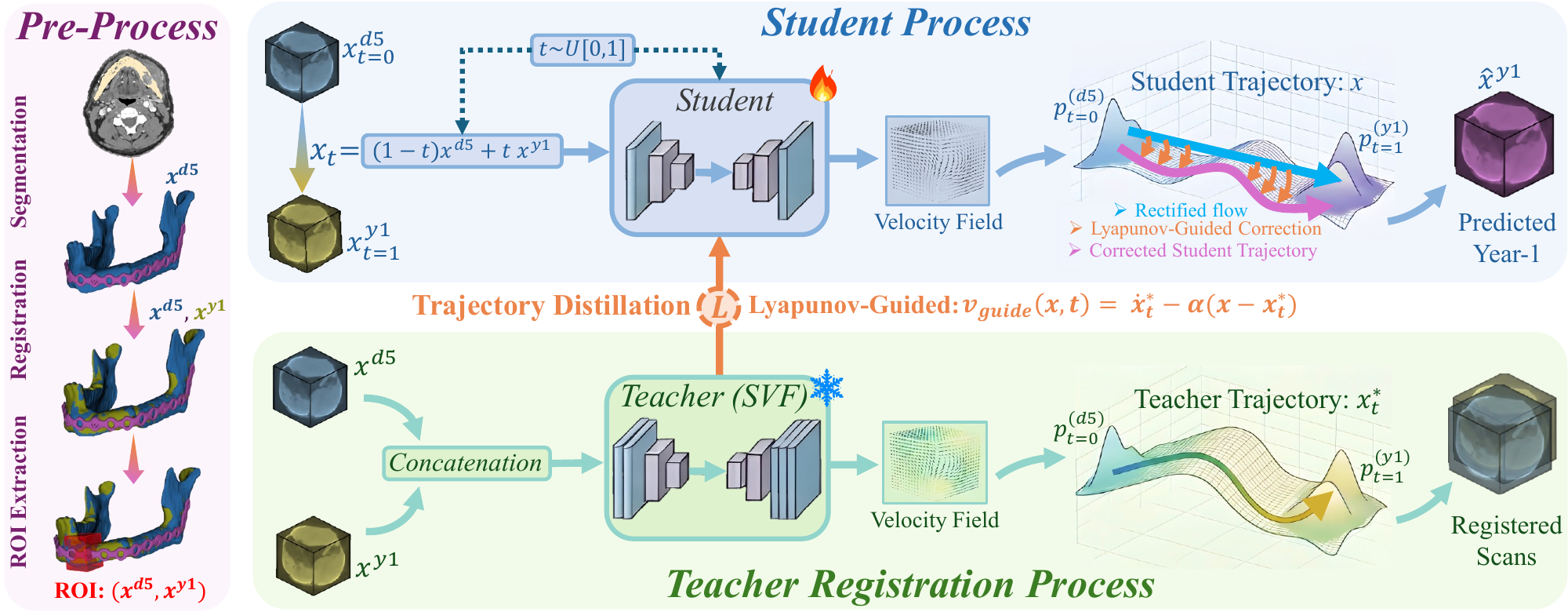} \caption{Illustration of the preprocessing steps and the teacher–student Lyapunov-guided distillation framework for guiding the student velocity field.} \label{fig:method} 
\end{figure}
In parallel, generative AI, including diffusion-based models and recent flow-matching formulations, has shown strong performance across medical imaging modalities, including anomaly detection~\cite{wyatt2022anoddpm}, modalities translation~\cite{xing2024cross} and fusion~\cite{zhou2025clinicalfmamba}, image synthesis~\cite{chang2025controllable,guo2025maisi}, and reconstruction~\cite{sun2024difr3ct}. These advances motivate a data-driven approach for modeling post-operative anatomical change directly from imaging. However, most existing medical image generation methods are not designed for longitudinal remodeling prediction, where the goal is not only realistic appearance but biologically plausible evolution from an early post-operative scan to a later healing state. Some approaches also rely on segmentation/mask-based conditioning~\cite{chu2024anatomic,konz2024anatomically}. While these constraints can improve anatomically consistent structure, they provide limited supervision for the transport dynamics governing longitudinal anatomical evolution. Moreover, prior distillation strategies in generative modeling commonly focus on one-step teacher matching~\cite{zhu2025simple,chang2025controllable}. This approach is not well suited to longitudinal transport settings, where the key challenge is trajectory-level consistency over the full evolution, beyond ODE solver or sampler improvements.

In this work, we formulate mandibular bone remodeling prediction as conditional generative transport between paired Day-5 and Year-1 CT regions of interest (ROIs) centered at the resection interface (Figure \ref{fig:method}). Our main contributions are as follows: (i) We develop a flow-based generative model, \textit{OsteoFlow}, that predicts a time-dependent image-space velocity field and integrates it over transport time to obtain the Year-1 prediction. (ii) To improve longitudinal transport learning, we cast teacher guidance as learning using privileged information (LUPI)~\cite{vapnik2009new}, where a registration-derived teacher computed from both scans is used as a prior to supervise the student trajectory. (iii) We introduce an effective yet simple control-inspired, Lyapunov-based trajectory distillation that goes beyond one-step teacher matching by distilling a \emph{continuous teacher trajectory} to improve rollout stability, anatomical coherence, and endpoint prediction. We further combine this trajectory-level supervision with rectified-flow supervision and a bone-focused image-space loss to improve endpoint fidelity near the union interface. Next, we describe our methodology in detail.

\section{Methods}
\begin{algorithm}[t]
\scriptsize
\caption{Flow-Based Trajectory Distillation with Lyapunov Guidance}
\label{alg:svf_teacher_compact}
\begin{algorithmic}[1]
\Require $(x^{d5},x^{y1})$; student $v_\theta$; teacher $v_{\mathrm{SVF}}$; $\alpha$; $\lambda_{\mathrm{img}}$; $\lambda_{\max}$; $\Delta t$; $\tau_{\mathrm{HU}}$; $\sigma$
\State \textbf{Training:} repeat \Comment{Optimize $v_\theta$}
\For{each iteration $k$ (sample $(x^{d5},x^{y1})$ and $t\sim\mathcal{U}(0,1)$)} \Comment{One update}
  \State $p_{\mathrm{on}}\gets \textsc{Ramp}(k)$, \quad $\lambda_{\mathrm{Lyap}}\gets \lambda_{\max}\cdot\textsc{Ramp}(k)$ \Comment{Joint ramps}
  \State $x_t^{\mathrm{anchor}}\gets (1-t)x^{d5}+t x^{y1}$ \Comment{RF anchor (off-policy)}
  \State $\{x_{\mathrm{roll}}(s)\}_{s\in[0,1]}\gets \textsc{Rollout}(v_\theta,x^{d5})$ \Comment{Entire rollout}
  \State $(x_t,\hat{x}^{y1}) \gets \begin{cases}
    (x_{\mathrm{roll}}(t),\,x_{\mathrm{roll}}(1)), & \text{with prob. } p_{\mathrm{on}} \\
    (x_t^{\mathrm{anchor}},\,x_{\mathrm{roll}}(1)), & \text{otherwise}
  \end{cases}$ \Comment{off-policy $\rightarrow$ on-policy}

  \State $x_t^\star\gets \mathcal{W}\!\big(x^{d5},\exp(t\cdot v_{\mathrm{SVF}})\big)$ \Comment{Teacher warp}
  \State $x_{t+\Delta t}^\star\gets \mathcal{W}\!\big(x^{d5},\exp((t+\Delta t)\cdot v_{\mathrm{SVF}})\big)$ \Comment{Next warp}
  \State $\dot{x}_t^\star\gets (x_{t+\Delta t}^\star-x_t^\star)/\Delta t$ \Comment{Teacher velocity (fwd diff.)}

  \State $v_{\text{pred}}\gets v_\theta(x_t,t;x^{d5})$ \Comment{Student velocity}
  \State $\mathcal{L}_{\mathrm{RF}}\gets \|v_{\text{pred}}-(x^{y1}-x^{d5})\|^2$ \Comment{RF supervision}
  \State $v_{\mathrm{guide}}\gets \dot{x}_t^\star-\alpha(x_t-x_t^\star)$ \Comment{Lyapunov guidance}
  \State $\mathcal{L}_{\mathrm{Lyap}}\gets \|v_{\text{pred}}-v_{\mathrm{guide}}\|^2$ \Comment{Trajectory distillation}

  \State $M_{\mathrm{bone}}(\omega)\gets \mathbf{1}\{x^{d5}(\omega)>\tau_{\mathrm{HU}}\}$ \Comment{Bone mask}
  \State $W(\omega)\propto \exp\!\big(-d(\omega)^2/(2\sigma^2)\big)\,M_{\mathrm{bone}}(\omega)$ \Comment{Resection-aware weight}
  \State $\mathcal{L}_{\mathrm{img}}\gets \|W\odot(\hat{x}^{y1}-x^{y1})\|^2$ \Comment{Bone-focused image loss}

  \State $\mathcal{L}\gets \mathcal{L}_{\mathrm{RF}}+\lambda_{\mathrm{img}}\mathcal{L}_{\mathrm{img}}+\lambda_{\mathrm{Lyap}}\mathcal{L}_{\mathrm{Lyap}}$ \Comment{Total loss}
  \State Update $\theta$ by minimizing $\mathcal{L}$ \Comment{Gradient step}
\EndFor
\State \textbf{Inference:} integrate $\dot{x}=v_\theta(x,t;x^{d5})$ from $t=0$ to $t=1$ \Comment{Teacher off}
\end{algorithmic}
\end{algorithm}
\textbf{Overview.} 
Let $(x^{d5},x^{y1}) \sim p_{\mathrm{data}}$ denote paired Day-5 and Year-1 post-operative regions of interest (ROIs), and let $t\sim\mathcal{U}(0,1)$ denote the transport time.
We model bone remodeling evolution by transporting $x^{d5}$ to $x^{y1}$ under a conditional velocity field $v_\theta(\cdot,t;x^{d5})$. The training objective has three terms: (i) a rectified-flow (RF) loss that provides paired supervision and learns the global remodeling transport direction, (ii) a Lyapunov-guided trajectory distillation loss that uses a registration-derived teacher trajectory as a LUPI-style prior to correct transport and improve endpoint accuracy, and (iii) a bone-focused image-space loss that improves endpoint fidelity in bone regions. Together, RF initializes and stabilizes transport, the teacher-guided term supervises the trajectory, and the image-space loss enforces Year-1 endpoint fidelity.

\textbf{Rectified-flow supervision.}
Training starts from the RF anchor state $x_t=(1-t)x^{d5}+t x^{y1}$, which defines the training states and initializes the model toward the endpoint direction $(x^{y1}-x^{d5})$. The RF loss learns the conditional velocity $v_\theta(x_t,t; x^{d5})$ toward the endpoint~\cite{liu2022flow}:
\begin{equation}
\mathcal{L}_{\mathrm{RF}}
=
\mathbb{E}_{p_{\mathrm{data}},\,t}
\!\left[
\big\|v_\theta(x_t,t;x^{d5})-(x^{y1}-x^{d5})\big\|_2^2
\right].
\label{eq:lrf}
\end{equation}
\begin{figure}[t]
\centering
\includegraphics[width=1\linewidth,height=.205\textheight]{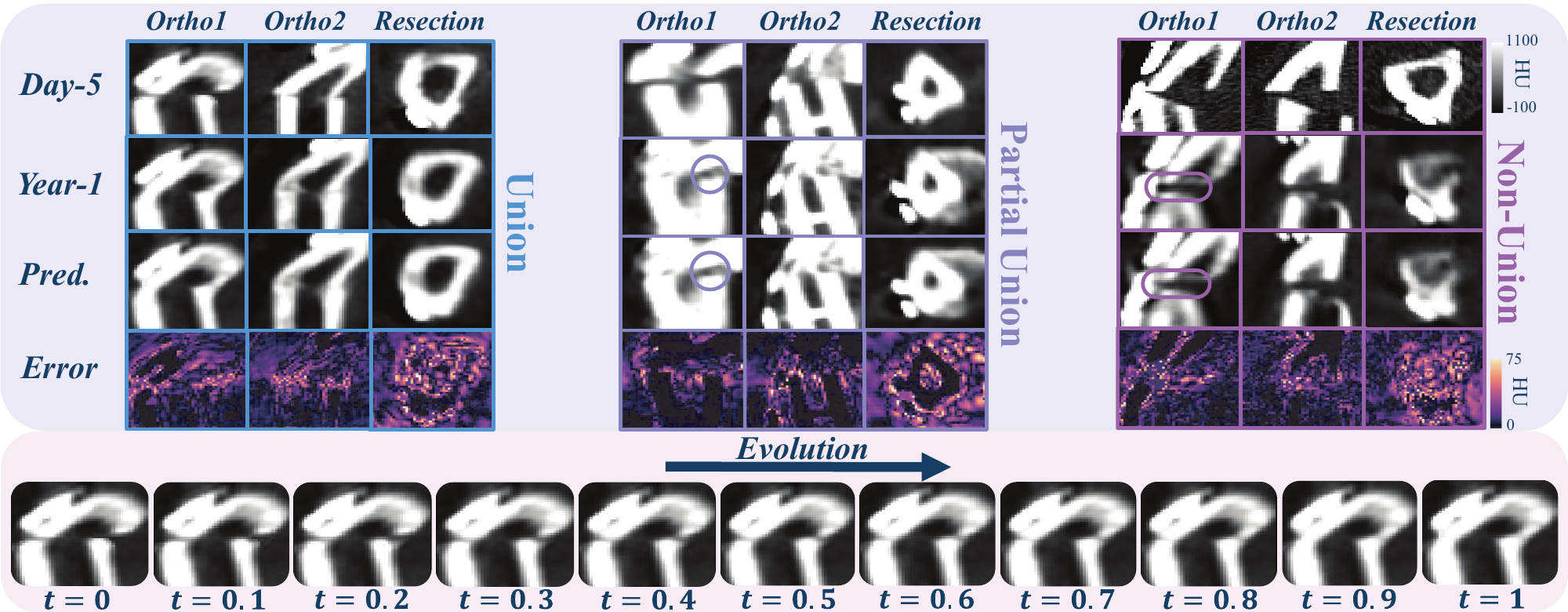}
\caption{Model predictions for three representative cases (union, partial union, and nonunion), shown on the resection plane and two other central orthogonal slices. \textcolor[HTML]{9186FE}{The purple circle} shows the partial union near the resection site, while the \textcolor[HTML]{C751FF}{pink ellipsoid} depicts the nonunion.
}
\label{fig:images}
\end{figure}\textbf{Lyapunov-guided trajectory distillation.}
During training, we gradually inject guidance from a teacher obtained via diffeomorphic stationary-velocity-field (SVF)~\cite{pai2015kernel} registration using privileged information from both scans; the teacher is used only during training and discarded at inference. It strictly provides deformation-based trajectory guidance as a \emph{prior} during training and does not model actual osteogenesis. Unlike common one-step distillation in generative AI, which matches a single teacher update, our method distills a \emph{continuous trajectory} over transport time. A simple baseline is mean-squared matching of student and teacher velocities, but this local objective does not enforce trajectory-level consistency in longitudinal transport. We therefore use a Lyapunov-guided formulation~\cite{khalil2002nonlinear} that imposes principled reference-tracking dynamics: the teacher path defines the guidance trajectory, and the Lyapunov term provides corrective feedback to keep the student close when local predictions are imperfect. We denote the teacher trajectory by $x^\star_t$, constructed from a teacher SVF $v_{\mathrm{SVF}}(\xi)$, and approximate its image-space velocity as
\begin{equation}
x_t^\star=\mathcal{W}\!\left(x^{\mathrm{d5}},\,\exp\!\left(t\cdot v_{\mathrm{SVF}}\right)\right)=x^{\mathrm{d5}}\circ\exp\!\left(t\cdot v_{\mathrm{SVF}}\right),
\qquad
\dot{x}_t^\star \approx \frac{x_{t+\Delta t}^\star - x_t^\star}{\Delta t}.
\label{eq:teacher_path_and_tangent}
\end{equation}
Since the SVF is defined in coordinate space (i.e., it parameterizes spatial deformation), it does not directly provide the image-space transport velocity needed for supervision. We therefore exponentiate the SVF using the scaling-and-squaring algorithm (warp operator $\mathcal{W}(\cdot,\cdot)$) at times $t$ and $t+\Delta t$, then apply the finite-difference approximation above to obtain the image-space velocity $\dot{x}_t^\star$. To distill this trajectory into the student dynamics, we define a quadratic Lyapunov energy centered on the teacher path,
\begin{equation}
V(x,t)=\frac{\alpha}{2}\,\|x-x_t^\star\|_2^2,
\qquad 
\nabla_x V(x,t)=\alpha\,(x-x_t^\star).
\label{eq:explicit_value}
\end{equation}
This Lyapunov function is positive semidefinite (zero on the reference path and positive elsewhere), and the feedback is designed so that its derivative along the closed-loop dynamics is non-increasing ($\dot{V} \leq 0$)~\cite{khalil2002nonlinear}. Let $u$ denote the control input. With running cost $\ell(x,t,u)=\frac{1}{2}\|u-\dot{x}_t^\star\|_2^2$, the induced guidance field is
$v_{\mathrm{guide}}(x,t)=\dot{x}_t^\star-\alpha\big(x-x_t^\star\big)$,
where $\dot{x}_t^\star$ is the teacher trajectory signal and $-\alpha(x-x_t^\star)$ is corrective feedback toward the guidance  path with stiffness $\alpha>0$. Unlike critic-based methods (e.g., actor-critic), this guidance is derived in closed form from a Lyapunov energy function, without requiring a separate critic network. We then distill this guidance into the student velocity field through
\begin{equation}
\mathcal{L}_{\mathrm{Lyap}}
=
\mathbb{E}_{p_{\mathrm{data}},\,t}
\!\left[
\big\|v_\theta(x_t,t;x^{d5})-v_{\mathrm{guide}}(x_t,t)\big\|_2^2
\right].
\label{eq:llyap}
\end{equation}
\textbf{Bone-focused image-space supervision.}
To improve endpoint fidelity in clinically relevant regions, we add a bone-focused image-space loss with a fixed, case-specific spatial weight map $W\in[0,1]$ that emphasizes bone voxels near the resection region. We define a Day-5 Hounsfield unit (HU)-derived bone mask as $M_{\mathrm{bone}}(\omega)=\mathbf{1}\!\{x^{d5}(\omega)>\tau_{\mathrm{HU}}=300\}$. Let $d(\omega)$ denote the distance from voxel $\omega$ to the resection plane, and let $\hat{x}^{y1}$ denote the predicted Year-1 image obtained by integrating the learned ODE from $x^{d5}$. We then define the resection-aware weight and image-space loss as
\begin{equation}
W(\omega)\propto \exp\!\left(-d(\omega)^2/(2\sigma^2)\right) M_{\mathrm{bone}}(\omega),
\quad
\mathcal{L}_{\mathrm{img}}
=
\mathbb{E}_{p_{\mathrm{data}}}
\!\left[
\big\|W\odot\big(\hat{x}^{y1}-x^{y1}\big)\big\|_2^2
\right].
\label{eq:bone_weight_and_img_loss}
\end{equation}
\textbf{Final objective.}
The student objective combines three losses:
\begin{equation}
\mathcal{L}=\mathcal{L}_{\mathrm{RF}}
+\lambda_{\mathrm{img}}\,\mathcal{L}_{\mathrm{img}}
+\lambda_{\mathrm{Lyap}}\,\mathcal{L}_{\mathrm{Lyap}},
\label{eq:combined_loss}
\end{equation}
where $\lambda_{\mathrm{img}}$ is fixed and $\lambda_{\mathrm{Lyap}}$ weights Lyapunov distillation. The teacher is trained only with $\mathcal{L}_{\mathrm{img}}$ and then frozen for student training. Since $\mathcal{L}_{\mathrm{RF}}$ and $\mathcal{L}_{\mathrm{Lyap}}$ can impose competing velocity supervision, we use a joint curriculum rather than fixed weighting (Algorithm~\ref{alg:svf_teacher_compact}). Training starts with RF supervision on off-policy anchor states for stability, then linearly increases both the Lyapunov-distillation weight and the probability of on-policy sampling. On-policy losses are evaluated on  rollout states, progressively shifting training toward inference-time dynamics.

\section{Experiments and Results}

\newcommand{\cmark}{\ding{51}}  
\newcommand{\xmark}{\ding{55}}  
\newcolumntype{C}[1]{>{\centering\arraybackslash}p{#1}}

\renewcommand{\cmark}{\scriptsize\checkmark}
\renewcommand{\xmark}{\scriptsize$\times$}

\renewcommand{\cmark}{\scriptsize\checkmark}
\renewcommand{\xmark}{\scriptsize$\times$}

\renewcommand{\cmark}{\scriptsize\checkmark}
\renewcommand{\xmark}{\scriptsize$\times$}
\subsection{Dataset and Augmentation Pipeline}
We curated an internal longitudinal mandibular CT dataset from 120 patients with scans at approximately postoperative Day-5 ($x^{d5}$) and Year-1 ($x^{y1}$). The mandible was segmented only to standardize landmarking. At each resection site, four fiducials were placed on $x^{d5}$ (where osteotomy margins are clear), while $x^{y1}$ often shows remodeling and bridging bone. The titanium reconstruction plate was segmented at both time points and used as a rigid reference to register $x^{y1}$ to $x^{d5}$. A plane fit to each 4-fiducial group defined a local frame, and an oblique cubic ROI centered on the resection site was resampled to $48^3$ voxels (0.5 mm isotropic), yielding 344 paired ROIs. All ROIs were windowed to $[-100, 1100]$ HU and affine-registered to suppress metal-driven outliers and focus on bone-remodeling-relevant intensities. The upper bound preserves cortical bone, and bridging callus near the resection typically falls below $\sim$1000 HU~\cite{datarkar2021qualitative}, leaving a 100 HU safety margin. To expand training data while preserving longitudinal correspondence, we used paired augmentation by applying the same transform to both ROIs in each $(x^{d5},x^{y1})$ pair. Geometric changes were limited to medically plausible perturbations: left-right mirroring, small rotations (up to $10^\circ$), and translations (up to 2 voxels). Pose augmentation was implemented by transforming the ROI sampling frame and re-extracting from the original volumes. Intensity augmentation included additive Gaussian noise (10--25 HU), mild brightness/contrast shifts (about $\pm$20--30 HU and $\pm$3\%), and light Gaussian blur ($\sigma=0.8$) to mimic reconstruction-kernel variability. Each ROI pair yielded 41 variants in total, resulting in 14,104 paired samples. We evaluated the model using three random patient-level train/test splits (85\% train, 15\% test per split). Augmentation was applied only to the training set, while the test sets used only original ROIs, preventing data leakage.
\subsection{Implementation Details and Metrics}
The student is a time-conditioned 3D residual U-Net~\cite{cciccek20163d} that predicts an image-space velocity field on $48^3$ ROIs (single-channel in/out). It uses two strided downsampling stages (48$\rightarrow$24$\rightarrow$12) with symmetric transposed-convolution upsampling, base width 48 channels (48/96/192 across scales), and residual blocks with $3\times3\times3$ convolutions, GroupNorm, and SiLU. Time is injected in every residual block via FiLM-style modulation (per-channel scale/shift)~\cite{perez2018film} after the first convolution, and the block output is added to the skip path. The teacher is an SVF 3D residual U-Net with the same backbone (base width 48; depth 48$\rightarrow$24$\rightarrow$12). It takes the concatenated pair $(x^{d5},x^{y1})$ as a 2-channel input and outputs a 3-channel SVF. Teacher parameters are frozen during student training, and to prevent data leakage, the teacher is retrained separately for each random split. In Eq.~\eqref{eq:combined_loss}, we set $\lambda_{\mathrm{img}}=0.2$ and linearly ramp $\lambda_{\mathrm{Lyap}}$ to $\lambda_{\max}=1$ over the first 10 epochs. In Eq.~\eqref{eq:explicit_value}, we set $\alpha=1$ empirically and leave sensitivity analysis for future work. Training used batch size 1 and AdamW ($10^{-4}$ learning rate, $10^{-5}$ weight decay) on a single A100 GPU (40 GB) for up to 50 epochs ($\sim600{,}000$ steps). Early stopping used a patience of 10 epochs based on validation plateau. Flow-based inference used RK4 ODE solver with step size 0.1. 

We compared \textit{OsteoFlow} against baselines spanning direct $(x^{d5},x^{y1})$ comparison, regression (ResUNet++~\cite{jha2019resunet++}), GANs (Pix2Pix~\cite{isola2017image}, GRIT~\cite{suri2024grit}), diffusions (cDDPM($\Delta$)~\cite{ho2020denoising,jiang2024ciresdiff}, SegGuidedDiff~\cite{konz2024anatomically}), VAEs (MedVAE~\cite{varma2025medvae}), and flow (Rectified Flow~\cite{liu2022flow}). Except for MedVAE, all baselines were matched to our backbone depth and capacity, differing only by method-specific architecture or losses. For MedVAE, we fine-tuned the released pretrained model with its original setup (L1 + LPIPS + PatchGAN; MedVAE-1); due to poor performance, we trained a second variant with the image-space loss in Eq.~\ref{eq:bone_weight_and_img_loss} (MedVAE-2). For cDDPM($\Delta$), following Jiang \textit{et al.}~\cite{jiang2024ciresdiff}, the model predicts the inter-scan intensity difference ($\Delta$) conditioned on $x^{d5}$. In SegGuidedDiff, we conditioned the model on $x^{d5}$ and the bony mask. We report Dice (bone, $HU>300$), multi-scale structural similarity (MS-SSIM), and mean absolute error (MAE; all HU and bone-only HU) for the full ROI and a mid slab (central 12 slices around the resection plane).
\begin{table}[t]
\centering
\scriptsize
\setlength{\tabcolsep}{1.8pt}
\renewcommand{\arraystretch}{.000001}
\setlength{\extrarowheight}{0pt}
\setcellgapes{-.5pt}
\makegapedcells
\caption{Baseline comparison (mean $\pm$ std across runs). Among non-teacher methods, the best and second-best results are shown in \textbf{bold} and \underline{underlined}, respectively. An asterisk (*) indicates a significant difference from the second-best result  using a two-sided Wilcoxon signed-rank test on paired ROI-level test metrics ($p<0.05$).}
\label{tab:baseline_comparison}

\begin{tabular}{@{}
>{\raggedright\arraybackslash}p{2.5cm}
cccc cccc
@{}}
\toprule

\multirow{2}{*}{Method} &
\multicolumn{4}{c}{Mid slab} &
\multicolumn{4}{c}{Full volume} \\
\cmidrule(lr){2-5}\cmidrule(lr){6-9}

&
\makecell{Dice\\(\%)$\uparrow$} &
\makecell{MS-SSIM\\(\%)$\uparrow$} &
\makecell{MAE\\All HU$\downarrow$} &
\makecell{MAE\\Bone HU$\downarrow$} &
\makecell{Dice\\(\%)$\uparrow$} &
\makecell{MS-SSIM\\(\%)$\uparrow$} &
\makecell{MAE\\All HU$\downarrow$} &
\makecell{MAE\\Bone HU$\downarrow$} \\
\midrule

Day5$\leftrightarrow$Year1 &
\makecell{85.60\\[-3.5pt]{\scalebox{0.8}{$\scriptscriptstyle \pm 0.18$}}} &
\makecell{64.18\\[-3.5pt]{\scalebox{0.8}{$\scriptscriptstyle \pm 0.36$}}} &
\makecell{95.89\\[-3.5pt]{\scalebox{0.8}{$\scriptscriptstyle \pm 2.93$}}} &
\makecell{177.94\\[-3.5pt]{\scalebox{0.8}{$\scriptscriptstyle \pm 5.40$}}} &
\makecell{94.36\\[-3.5pt]{\scalebox{0.8}{$\scriptscriptstyle \pm 0.12$}}} &
\makecell{87.13\\[-3.5pt]{\scalebox{0.8}{$\scriptscriptstyle \pm 0.26$}}} &
\makecell{41.78\\[-3.5pt]{\scalebox{0.8}{$\scriptscriptstyle \pm 1.75$}}} &
\makecell{70.32\\[-3.5pt]{\scalebox{0.8}{$\scriptscriptstyle \pm 3.27$}}} \\

ResUNet++~\cite{jha2019resunet++} &
\makecell{92.65\\[-3.5pt]{\scalebox{0.8}{$\scriptscriptstyle \pm 0.11$}}} &
\makecell{78.10\\[-3.5pt]{\scalebox{0.8}{$\scriptscriptstyle \pm 0.22$}}} &
\makecell{56.85\\[-3.5pt]{\scalebox{0.8}{$\scriptscriptstyle \pm 1.64$}}} &
\makecell{102.67\\[-3.5pt]{\scalebox{0.8}{$\scriptscriptstyle \pm 4.20$}}} &
\makecell{96.57\\[-3.5pt]{\scalebox{0.8}{$\scriptscriptstyle \pm 0.08$}}} &
\makecell{91.82\\[-3.5pt]{\scalebox{0.8}{$\scriptscriptstyle \pm 0.17$}}} &
\makecell{27.76\\[-3.5pt]{\scalebox{0.8}{$\scriptscriptstyle \pm 1.12$}}} &
\makecell{45.80\\[-3.5pt]{\scalebox{0.8}{$\scriptscriptstyle \pm 2.50$}}} \\

Pix2Pix~\cite{isola2017image}&
\makecell{92.51\\[-3.5pt]{\scalebox{0.8}{$\scriptscriptstyle \pm 0.11$}}} &
\makecell{78.26\\[-3.5pt]{\scalebox{0.8}{$\scriptscriptstyle \pm 0.25$}}} &
\makecell{56.96\\[-3.5pt]{\scalebox{0.8}{$\scriptscriptstyle \pm 1.80$}}} &
\makecell{103.55\\[-3.5pt]{\scalebox{0.8}{$\scriptscriptstyle \pm 4.50$}}} &
\makecell{96.45\\[-3.5pt]{\scalebox{0.8}{$\scriptscriptstyle \pm 0.08$}}} &
\makecell{91.86\\[-3.5pt]{\scalebox{0.8}{$\scriptscriptstyle \pm 0.18$}}} &
\makecell{27.75\\[-3.5pt]{\scalebox{0.8}{$\scriptscriptstyle \pm 1.26$}}} &
\makecell{46.38\\[-3.5pt]{\scalebox{0.8}{$\scriptscriptstyle \pm 2.71$}}} \\

GRIT~\cite{suri2024grit}&
\makecell{92.45\\[-3.5pt]{\scalebox{0.8}{$\scriptscriptstyle \pm 0.13$}}} &
\makecell{77.05\\[-3.5pt]{\scalebox{0.8}{$\scriptscriptstyle \pm 0.31$}}} &
\makecell{60.07\\[-3.5pt]{\scalebox{0.8}{$\scriptscriptstyle \pm 2.21$}}} &
\makecell{100.63\\[-3.5pt]{\scalebox{0.8}{$\scriptscriptstyle \pm 4.10$}}} &
\makecell{96.48\\[-3.5pt]{\scalebox{0.8}{$\scriptscriptstyle \pm 0.09$}}} &
\makecell{91.32\\[-3.5pt]{\scalebox{0.8}{$\scriptscriptstyle \pm 0.22$}}} &
\makecell{30.31\\[-3.5pt]{\scalebox{0.8}{$\scriptscriptstyle \pm 1.56$}}} &
\makecell{45.78\\[-3.5pt]{\scalebox{0.8}{$\scriptscriptstyle \pm 2.63$}}} \\

cDDPM($\Delta$)~\cite{ho2020denoising,jiang2024ciresdiff} &
\makecell{\underline{93.64}\\[-3pt]{\scalebox{0.8}{$\scriptscriptstyle \pm 0.15$}}} &
\makecell{\underline{80.02}\\[-3pt]{\scalebox{0.8}{$\scriptscriptstyle \pm 0.34$}}} &
\makecell{\underline{50.45}\\[-3pt]{\scalebox{0.8}{$\scriptscriptstyle \pm 2.44$}}} &
\makecell{\underline{87.15}\\[-3pt]{\scalebox{0.8}{$\scriptscriptstyle \pm 4.81$}}} &
\makecell{\underline{97.01}\\[-3pt]{\scalebox{0.8}{$\scriptscriptstyle \pm 0.11$}}} &
\makecell{92.21\\[-3pt]{\scalebox{0.8}{$\scriptscriptstyle \pm 0.24$}}} &
\makecell{\underline{26.14}\\[-3pt]{\scalebox{0.8}{$\scriptscriptstyle \pm 1.62$}}} &
\makecell{\underline{39.71}\\[-3pt]{\scalebox{0.8}{$\scriptscriptstyle \pm 3.19$}}} \\

SegGuidedDiff~\cite{konz2024anatomically} &
\makecell{93.03\\[-3.5pt]{\scalebox{0.8}{$\scriptscriptstyle \pm 0.14$}}} &
\makecell{78.00\\[-3.5pt]{\scalebox{0.8}{$\scriptscriptstyle \pm 0.29$}}} &
\makecell{54.10\\[-3.5pt]{\scalebox{0.8}{$\scriptscriptstyle \pm 2.10$}}} &
\makecell{95.58\\[-3.5pt]{\scalebox{0.8}{$\scriptscriptstyle \pm 4.60$}}} &
\makecell{96.93\\[-3.5pt]{\scalebox{0.8}{$\scriptscriptstyle \pm 0.10$}}} &
\makecell{91.57\\[-3pt]{\scalebox{0.8}{$\scriptscriptstyle \pm 0.23$}}} &
\makecell{27.40\\[-3.5pt]{\scalebox{0.8}{$\scriptscriptstyle \pm 1.40$}}} &
\makecell{43.06\\[-3.5pt]{\scalebox{0.8}{$\scriptscriptstyle \pm 2.88$}}} \\

MedVAE-1~\cite{varma2025medvae} &
\makecell{88.86\\[-3.5pt]{\scalebox{0.8}{$\scriptscriptstyle \pm 0.19$}}} &
\makecell{63.94\\[-3.5pt]{\scalebox{0.8}{$\scriptscriptstyle \pm 0.44$}}} &
\makecell{81.76\\[-3.5pt]{\scalebox{0.8}{$\scriptscriptstyle \pm 3.10$}}} &
\makecell{145.39\\[-3.5pt]{\scalebox{0.8}{$\scriptscriptstyle \pm 5.90$}}} &
\makecell{92.41\\[-3.5pt]{\scalebox{0.8}{$\scriptscriptstyle \pm 0.14$}}} &
\makecell{73.66\\[-3.5pt]{\scalebox{0.8}{$\scriptscriptstyle \pm 0.35$}}} &
\makecell{57.99\\[-3.5pt]{\scalebox{0.8}{$\scriptscriptstyle \pm 2.22$}}} &
\makecell{96.00\\[-3.5pt]{\scalebox{0.8}{$\scriptscriptstyle \pm 3.80$}}} \\

MedVAE-2~\cite{varma2025medvae} &
\makecell{89.63\\[-3.5pt]{\scalebox{0.8}{$\scriptscriptstyle \pm 0.17$}}} &
\makecell{67.56\\[-3.5pt]{\scalebox{0.8}{$\scriptscriptstyle \pm 0.39$}}} &
\makecell{73.78\\[-3.5pt]{\scalebox{0.8}{$\scriptscriptstyle \pm 2.74$}}} &
\makecell{107.29\\[-3.5pt]{\scalebox{0.8}{$\scriptscriptstyle \pm 5.40$}}} &
\makecell{92.80\\[-3.5pt]{\scalebox{0.8}{$\scriptscriptstyle \pm 0.12$}}} &
\makecell{76.67\\[-3.5pt]{\scalebox{0.8}{$\scriptscriptstyle \pm 0.31$}}} &
\makecell{51.12\\[-3.5pt]{\scalebox{0.8}{$\scriptscriptstyle \pm 2.00$}}} &
\makecell{66.98\\[-3.5pt]{\scalebox{0.8}{$\scriptscriptstyle \pm 3.50$}}} \\

RecFlow~\cite{liu2022flow} &
\makecell{92.86\\[-3.5pt]{\scalebox{0.8}{$\scriptscriptstyle \pm 0.13$}}} &
\makecell{79.25\\[-3.5pt]{\scalebox{0.8}{$\scriptscriptstyle \pm 0.28$}}} &
\makecell{55.24\\[-3.5pt]{\scalebox{0.8}{$\scriptscriptstyle \pm 2.05$}}} &
\makecell{97.28\\[-3.5pt]{\scalebox{0.8}{$\scriptscriptstyle \pm 4.45$}}} &
\makecell{96.70\\[-3.5pt]{\scalebox{0.8}{$\scriptscriptstyle \pm 0.12$}}} &
\makecell{\underline{92.32}\\[-3pt]{\scalebox{0.8}{$\scriptscriptstyle \pm 0.22$}}} &
\makecell{26.98\\[-3.5pt]{\scalebox{0.8}{$\scriptscriptstyle \pm 1.54$}}} &
\makecell{45.06\\[-3.5pt]{\scalebox{0.8}{$\scriptscriptstyle \pm 3.04$}}} \\

\midrule

\textbf{\textit{OsteoFlow}~(Ours)} &
\makecell{\textbf{94.55}$^{*}$\\[-3.5pt]{\scalebox{0.8}{\boldmath$\scriptscriptstyle \pm 0.09$}}} &
\makecell{\textbf{83.49}$^{*}$\\[-3.5pt]{\scalebox{0.8}{\boldmath$\scriptscriptstyle \pm 0.21$}}} &
\makecell{\textbf{44.12}$^{*}$\\[-3.5pt]{\scalebox{0.8}{\boldmath$\scriptscriptstyle \pm 1.34$}}} &
\makecell{\textbf{69.88}$^{*}$\\[-3.5pt]{\scalebox{0.8}{\boldmath$\scriptscriptstyle \pm 3.75$}}} &
\makecell{\textbf{97.23}$^{*}$\\[-3.5pt]{\scalebox{0.8}{\boldmath$\scriptscriptstyle \pm 0.07$}}} &
\makecell{\textbf{93.82}$^{*}$\\[-3.5pt]{\scalebox{0.8}{\boldmath$\scriptscriptstyle \pm 0.15$}}} &
\makecell{\textbf{22.88}$^{*}$\\[-3.5pt]{\scalebox{0.8}{\boldmath$\scriptscriptstyle \pm 0.92$}}} &
\makecell{\textbf{36.74}$^{*}$\\[-3.5pt]{\scalebox{0.8}{\boldmath$\scriptscriptstyle \pm 2.30$}}} \\

Teacher~(LUPI) &
\makecell{96.87\\[-3.5pt]{\scalebox{0.8}{$\scriptscriptstyle \pm 0.07$}}} &
\makecell{90.33\\[-3.5pt]{\scalebox{0.8}{$\scriptscriptstyle \pm 0.16$}}} &
\makecell{37.17\\[-3.5pt]{\scalebox{0.8}{$\scriptscriptstyle \pm 1.09$}}} &
\makecell{40.63\\[-3.5pt]{\scalebox{0.8}{$\scriptscriptstyle \pm 3.28$}}} &
\makecell{98.45\\[-3.5pt]{\scalebox{0.8}{$\scriptscriptstyle \pm 0.05$}}} &
\makecell{96.55\\[-3.5pt]{\scalebox{0.8}{$\scriptscriptstyle \pm 0.12$}}} &
\makecell{19.31\\[-3.5pt]{\scalebox{0.8}{$\scriptscriptstyle \pm 0.76$}}} &
\makecell{18.43\\[-3.5pt]{\scalebox{0.8}{$\scriptscriptstyle \pm 1.98$}}} \\

\bottomrule
\end{tabular}
\end{table}
\subsection {Results and Ablation Studies}
Table~\ref{tab:baseline_comparison} summarizes the comparison between our method (\textit{OsteoFlow}) and state-of-the-art baselines. \textit{OsteoFlow} achieves the best overall performance with the most important gains in the mid-slab region, which contains the graft-host zone and the main remodeling changes. Relative to the second best method (cDDPM($\Delta$)), \textit{OsteoFlow} improves mid-slab Dice from 93.64 to 94.55 (+0.91 points, $\sim$0.97\%), improves MS-SSIM from 80.02 to 83.49 (+3.47 points, $\sim$4.3\%), and reduces MAE (All HU) from 50.45 to 44.12 HU ($-6.33$ HU, $\sim$12.5\%). The largest gain is in mid-slab bone MAE, which drops from 87.15 to 69.88 HU ($-17.27$ HU, $\sim$19.8\%), indicating markedly better fidelity of bone remodeling near the union site. These gains are notable because they occur in a high-performance regime (e.g., mid-slab Dice $>93\%$), where further improvements are typically harder to obtain. As expected, the teacher, constructed using privileged information from both scans, acts as a structural upper bound. While the teacher is limited to spatial deformation for guiding the student's trajectory and cannot predict true osteogenesis, \textit{OsteoFlow} narrows this performance gap. Figure~\ref{fig:images} also shows plausible union, nonunion, and partial union prediction, rather than smooth, averaged outputs, despite union being the dominant class.
\begin{table}[t]
\centering
\scriptsize
\setlength{\tabcolsep}{1.74pt}
\renewcommand{\arraystretch}{1.0}
\setlength{\extrarowheight}{0pt}
\setcellgapes{-.5pt}
\makegapedcells

\caption{Ablation study (mean $\pm$ std across runs). $R$ = rectified flow; $T_1$ = our Lyapunov-guided distillation; $T_2$ = MSE-based distillation; $A$ = augmentation; $I$ = image-space loss; $W$ = resection-aware weight. Best results are in \textbf{bold}.}
\label{tab:ablation_mean_percent}
\begin{tabular}{@{}
>{\centering\scriptsize}p{0.3cm}
>{\centering\scriptsize}p{0.3cm}
>{\centering\scriptsize}p{0.3cm}
>{\centering\scriptsize}p{0.3cm}
>{\centering\scriptsize}p{0.3cm}
>{\centering\scriptsize}p{0.3cm}
>{\centering\scriptsize}p{0.3cm}
cccc cccc
@{}}
\toprule
&
\multirow{2}{*}{$R$} &
\multirow{2}{*}{$A$} &
\multirow{2}{*}{$I$} &
\multirow{2}{*}{$T_1$} &
\multirow{2}{*}{$T_2$} &
\multirow{2}{*}{$W$} &
\multicolumn{4}{c}{Mid slab} &
\multicolumn{4}{c}{Full volume} \\
\cmidrule(lr){8-11}\cmidrule(lr){12-15}

& & & & & & &
\makecell{Dice\\(\%)$\uparrow$} &
\makecell{MS-SSIM\\(\%)$\uparrow$} &
\makecell{MAE\\All HU$\downarrow$} &
\makecell{MAE\\Bone HU$\downarrow$} &
\makecell{Dice\\(\%)$\uparrow$} &
\makecell{MS-SSIM\\(\%)$\uparrow$} &
\makecell{MAE\\All HU$\downarrow$} &
\makecell{MAE\\Bone HU$\downarrow$} \\
\midrule

\multirow{4}{*}{\rotatebox{90}{\hspace{-.35cm}\tiny{Teacher Effect}}} &
\cmark & \xmark & \xmark & \xmark & \xmark & \xmark &
\makecell{88.97\\[-3.5pt]{\scalebox{0.8}{$\scriptscriptstyle \pm 0.20$}}} &
\makecell{66.76\\[-3.5pt]{\scalebox{0.8}{$\scriptscriptstyle \pm 0.48$}}} &
\makecell{82.47\\[-3.5pt]{\scalebox{0.8}{$\scriptscriptstyle \pm 3.08$}}} &
\makecell{139.84\\[-3.5pt]{\scalebox{0.8}{$\scriptscriptstyle \pm 5.86$}}} &
\makecell{95.25\\[-3.5pt]{\scalebox{0.8}{$\scriptscriptstyle \pm 0.13$}}} &
\makecell{87.45\\[-3.5pt]{\scalebox{0.8}{$\scriptscriptstyle \pm 0.30$}}} &
\makecell{40.07\\[-3.5pt]{\scalebox{0.8}{$\scriptscriptstyle \pm 1.92$}}} &
\makecell{63.89\\[-3.5pt]{\scalebox{0.8}{$\scriptscriptstyle \pm 3.96$}}} \\

& \xmark & \xmark & \xmark & \cmark & \xmark & \xmark &
\makecell{90.79\\[-3.5pt]{\scalebox{0.8}{$\scriptscriptstyle \pm 0.17$}}} &
\makecell{71.23\\[-3.5pt]{\scalebox{0.8}{$\scriptscriptstyle \pm 0.41$}}} &
\makecell{77.00\\[-3.5pt]{\scalebox{0.8}{$\scriptscriptstyle \pm 2.74$}}} &
\makecell{123.85\\[-3.5pt]{\scalebox{0.8}{$\scriptscriptstyle \pm 5.33$}}} &
\makecell{95.72\\[-3.5pt]{\scalebox{0.8}{$\scriptscriptstyle \pm 0.11$}}} &
\makecell{89.05\\[-3.5pt]{\scalebox{0.8}{$\scriptscriptstyle \pm 0.27$}}} &
\makecell{37.08\\[-3.5pt]{\scalebox{0.8}{$\scriptscriptstyle \pm 1.70$}}} &
\makecell{55.91\\[-3.5pt]{\scalebox{0.8}{$\scriptscriptstyle \pm 3.58$}}} \\

& \cmark & \xmark & \xmark & \cmark & \xmark & \xmark &
\makecell{\textbf{91.54}\\[-3.5pt]{\scalebox{0.8}{\boldmath$\scriptscriptstyle \pm 0.16$}}} &
\makecell{\textbf{73.20}\\[-3.5pt]{\scalebox{0.8}{\boldmath$\scriptscriptstyle \pm 0.39$}}} &
\makecell{\textbf{69.79}\\[-3.5pt]{\scalebox{0.8}{\boldmath$\scriptscriptstyle \pm 2.41$}}} &
\makecell{\textbf{115.66}\\[-3.5pt]{\scalebox{0.8}{\boldmath$\scriptscriptstyle \pm 5.02$}}} &
\makecell{\textbf{96.00}\\[-3.5pt]{\scalebox{0.8}{\boldmath$\scriptscriptstyle \pm 0.10$}}} &
\makecell{\textbf{89.68}\\[-3.5pt]{\scalebox{0.8}{\boldmath$\scriptscriptstyle \pm 0.26$}}} &
\makecell{\textbf{34.51}\\[-3.5pt]{\scalebox{0.8}{\boldmath$\scriptscriptstyle \pm 1.49$}}} &
\makecell{\textbf{52.91}\\[-3.5pt]{\scalebox{0.8}{\boldmath$\scriptscriptstyle \pm 3.37$}}} \\

& \cmark & \xmark & \xmark & \xmark & \cmark & \xmark &
\makecell{89.67\\[-3.5pt]{\scalebox{0.8}{$\scriptscriptstyle \pm 0.22$}}} &
\makecell{69.22\\[-3.5pt]{\scalebox{0.8}{$\scriptscriptstyle \pm 0.50$}}} &
\makecell{79.29\\[-3.5pt]{\scalebox{0.8}{$\scriptscriptstyle \pm 2.96$}}} &
\makecell{137.79\\[-3.5pt]{\scalebox{0.8}{$\scriptscriptstyle \pm 5.71$}}} &
\makecell{95.49\\[-3.5pt]{\scalebox{0.8}{$\scriptscriptstyle \pm 0.14$}}} &
\makecell{88.48\\[-3.5pt]{\scalebox{0.8}{$\scriptscriptstyle \pm 0.33$}}} &
\makecell{37.60\\[-3.5pt]{\scalebox{0.8}{$\scriptscriptstyle \pm 1.83$}}} &
\makecell{60.03\\[-3.5pt]{\scalebox{0.8}{$\scriptscriptstyle \pm 3.80$}}} \\

\midrule

\multirow{4}{*}{\rotatebox{90}{\hspace{-.5cm}\tiny{Full Pipeline}}} &
\cmark & \cmark & \xmark & \xmark & \xmark & \xmark &
\makecell{92.86\\[-3.5pt]{\scalebox{0.8}{$\scriptscriptstyle \pm 0.13$}}} &
\makecell{79.25\\[-3.5pt]{\scalebox{0.8}{$\scriptscriptstyle \pm 0.28$}}} &
\makecell{55.24\\[-3.5pt]{\scalebox{0.8}{$\scriptscriptstyle \pm 2.05$}}} &
\makecell{97.28\\[-3.5pt]{\scalebox{0.8}{$\scriptscriptstyle \pm 4.45$}}} &
\makecell{96.70\\[-3.5pt]{\scalebox{0.8}{$\scriptscriptstyle \pm 0.12$}}} &
\makecell{92.32\\[-3.5pt]{\scalebox{0.8}{$\scriptscriptstyle \pm 0.22$}}} &
\makecell{26.98\\[-3.5pt]{\scalebox{0.8}{$\scriptscriptstyle \pm 1.54$}}} &
\makecell{45.06\\[-3.5pt]{\scalebox{0.8}{$\scriptscriptstyle \pm 3.04$}}} \\

& \cmark & \cmark & \cmark & \xmark & \xmark & \xmark &
\makecell{93.65\\[-3.5pt]{\scalebox{0.8}{$\scriptscriptstyle \pm 0.12$}}} &
\makecell{81.39\\[-3.5pt]{\scalebox{0.8}{$\scriptscriptstyle \pm 0.25$}}} &
\makecell{49.47\\[-3.5pt]{\scalebox{0.8}{$\scriptscriptstyle \pm 1.74$}}} &
\makecell{83.65\\[-3.5pt]{\scalebox{0.8}{$\scriptscriptstyle \pm 4.06$}}} &
\makecell{97.00\\[-3.5pt]{\scalebox{0.8}{$\scriptscriptstyle \pm 0.10$}}} &
\makecell{93.09\\[-3.5pt]{\scalebox{0.8}{$\scriptscriptstyle \pm 0.19$}}} &
\makecell{24.95\\[-3.5pt]{\scalebox{0.8}{$\scriptscriptstyle \pm 1.26$}}} &
\makecell{40.02\\[-3.5pt]{\scalebox{0.8}{$\scriptscriptstyle \pm 2.92$}}} \\

& \cmark & \cmark & \cmark & \cmark & \xmark & \xmark &
\makecell{94.46\\[-3.5pt]{\scalebox{0.8}{$\scriptscriptstyle \pm 0.10$}}} &
\makecell{83.12\\[-3.5pt]{\scalebox{0.8}{$\scriptscriptstyle \pm 0.22$}}} &
\makecell{45.37\\[-3.5pt]{\scalebox{0.8}{$\scriptscriptstyle \pm 1.41$}}} &
\makecell{71.85\\[-3.5pt]{\scalebox{0.8}{$\scriptscriptstyle \pm 3.79$}}} &
\makecell{97.21\\[-3.5pt]{\scalebox{0.8}{$\scriptscriptstyle \pm 0.08$}}} &
\makecell{93.73\\[-3.5pt]{\scalebox{0.8}{$\scriptscriptstyle \pm 0.17$}}} &
\makecell{23.10\\[-3.5pt]{\scalebox{0.8}{$\scriptscriptstyle \pm 1.02$}}} &
\makecell{37.12\\[-3.5pt]{\scalebox{0.8}{$\scriptscriptstyle \pm 2.49$}}} \\

& \cmark & \cmark & \cmark & \cmark & \xmark & \cmark &
\makecell{\textbf{94.55}\\[-3.5pt]{\scalebox{0.8}{\boldmath$\scriptscriptstyle \pm 0.09$}}} &
\makecell{\textbf{83.49}\\[-3.5pt]{\scalebox{0.8}{\boldmath$\scriptscriptstyle \pm 0.21$}}} &
\makecell{\textbf{44.12}\\[-3.5pt]{\scalebox{0.8}{\boldmath$\scriptscriptstyle \pm 1.34$}}} &
\makecell{\textbf{69.88}\\[-3.5pt]{\scalebox{0.8}{\boldmath$\scriptscriptstyle \pm 3.75$}}} &
\makecell{\textbf{97.23}\\[-3.5pt]{\scalebox{0.8}{\boldmath$\scriptscriptstyle \pm 0.07$}}} &
\makecell{\textbf{93.82}\\[-3.5pt]{\scalebox{0.8}{\boldmath$\scriptscriptstyle \pm 0.15$}}} &
\makecell{\textbf{22.88}\\[-3.5pt]{\scalebox{0.8}{\boldmath$\scriptscriptstyle \pm 0.92$}}} &
\makecell{\textbf{36.74}\\[-3.5pt]{\scalebox{0.8}{\boldmath$\scriptscriptstyle \pm 2.30$}}} \\
\bottomrule
\end{tabular}
\end{table}
\begin{table}[t]
\centering
\scriptsize
\setlength{\tabcolsep}{2.5pt}
\renewcommand{\arraystretch}{1.0}
\setlength{\extrarowheight}{0pt}
\setcellgapes{-.5pt}
\makegapedcells

\caption{Backbone ablation study (mean $\pm$ std across runs). $A$ = augmentation; $I$ = image-space loss. Both backbones have $\sim$12M parameters and the same network depth for a fair comparison. Best results are in \textbf{bold}.}
\label{tab:backbone_ablation}

\begin{tabular}{@{}
>{\centering}p{1.6cm}
>{\centering\scriptsize}p{0.3cm}
>{\centering\scriptsize}p{0.3cm}
cccc cccc
@{}}
\toprule

\multirow{2}{*}{Backbone} &
\multirow{2}{*}{$A$} &
\multirow{2}{*}{$I$} &
\multicolumn{4}{c}{Mid slab} &
\multicolumn{4}{c}{Full volume} \\
\cmidrule(lr){4-7}\cmidrule(lr){8-11}

& & &
\makecell{Dice\\(\%)$\uparrow$} &
\makecell{MS-SSIM\\(\%)$\uparrow$} &
\makecell{MAE\\All HU$\downarrow$} &
\makecell{MAE\\Bone HU$\downarrow$} &
\makecell{Dice\\(\%)$\uparrow$} &
\makecell{MS-SSIM\\(\%)$\uparrow$} &
\makecell{MAE\\All HU$\downarrow$} &
\makecell{MAE\\Bone HU$\downarrow$} \\
\midrule

SwinUNETR & \cmark & \xmark &
\makecell{92.30\\[-3.5pt]{\scalebox{0.8}{$\scriptscriptstyle \pm 0.14$}}} &
\makecell{77.79\\[-3.5pt]{\scalebox{0.8}{$\scriptscriptstyle \pm 0.30$}}} &
\makecell{58.83\\[-3.5pt]{\scalebox{0.8}{$\scriptscriptstyle \pm 1.86$}}} &
\makecell{103.92\\[-3.5pt]{\scalebox{0.8}{$\scriptscriptstyle \pm 4.62$}}} &
\makecell{96.58\\[-3.5pt]{\scalebox{0.8}{$\scriptscriptstyle \pm 0.10$}}} &
\makecell{91.44\\[-3.5pt]{\scalebox{0.8}{$\scriptscriptstyle \pm 0.21$}}} &
\makecell{28.28\\[-3.5pt]{\scalebox{0.8}{$\scriptscriptstyle \pm 1.30$}}} &
\makecell{46.82\\[-3.5pt]{\scalebox{0.8}{$\scriptscriptstyle \pm 2.82$}}} \\

\textbf{Res-UNet} & \textbf{\cmark} & \textbf{\xmark} &
\makecell{\textbf{92.86}\\[-3.5pt]{\scalebox{0.8}{\boldmath$\scriptscriptstyle \pm 0.13$}}} &
\makecell{\textbf{79.25}\\[-3.5pt]{\scalebox{0.8}{\boldmath$\scriptscriptstyle \pm 0.28$}}} &
\makecell{\textbf{55.24}\\[-3.5pt]{\scalebox{0.8}{\boldmath$\scriptscriptstyle \pm 2.05$}}} &
\makecell{\textbf{97.28}\\[-3.5pt]{\scalebox{0.8}{\boldmath$\scriptscriptstyle \pm 4.45$}}} &
\makecell{\textbf{96.70}\\[-3.5pt]{\scalebox{0.8}{\boldmath$\scriptscriptstyle \pm 0.12$}}} &
\makecell{\textbf{92.32}\\[-3.5pt]{\scalebox{0.8}{\boldmath$\scriptscriptstyle \pm 0.22$}}} &
\makecell{\textbf{26.98}\\[-3.5pt]{\scalebox{0.8}{\boldmath$\scriptscriptstyle \pm 1.54$}}} &
\makecell{\textbf{45.06}\\[-3.5pt]{\scalebox{0.8}{\boldmath$\scriptscriptstyle \pm 3.04$}}} \\

\midrule

SwinUNETR & \cmark & \cmark &
\makecell{93.27\\[-3.2pt]{\scalebox{0.8}{$\scriptscriptstyle \pm 0.13$}}} &
\makecell{80.24\\[-3.2pt]{\scalebox{0.8}{$\scriptscriptstyle \pm 0.27$}}} &
\makecell{51.95\\[-3.2pt]{\scalebox{0.8}{$\scriptscriptstyle \pm 1.78$}}} &
\makecell{89.21\\[-3.2pt]{\scalebox{0.8}{$\scriptscriptstyle \pm 4.36$}}} &
\makecell{96.86\\[-3.2pt]{\scalebox{0.8}{$\scriptscriptstyle \pm 0.10$}}} &
\makecell{92.74\\[-3.2pt]{\scalebox{0.8}{$\scriptscriptstyle \pm 0.20$}}} &
\makecell{25.49\\[-3.2pt]{\scalebox{0.8}{$\scriptscriptstyle \pm 1.22$}}} &
\makecell{41.24\\[-3.2pt]{\scalebox{0.8}{$\scriptscriptstyle \pm 2.94$}}} \\
\textbf{Res-UNet} & \textbf{\cmark} & \textbf{\cmark} &
\makecell{\textbf{93.65}\\[-3.2pt]{\scalebox{0.8}{\boldmath$\scriptscriptstyle \pm 0.12$}}} &
\makecell{\textbf{81.39}\\[-3.2pt]{\scalebox{0.8}{\boldmath$\scriptscriptstyle \pm 0.25$}}} &
\makecell{\textbf{49.47}\\[-3.2pt]{\scalebox{0.8}{\boldmath$\scriptscriptstyle \pm 1.74$}}} &
\makecell{\textbf{83.65}\\[-3.2pt]{\scalebox{0.8}{\boldmath$\scriptscriptstyle \pm 4.06$}}} &
\makecell{\textbf{97.00}\\[-3.2pt]{\scalebox{0.8}{\boldmath$\scriptscriptstyle \pm 0.10$}}} &
\makecell{\textbf{93.09}\\[-3.2pt]{\scalebox{0.8}{\boldmath$\scriptscriptstyle \pm 0.19$}}} &
\makecell{\textbf{24.95}\\[-3.2pt]{\scalebox{0.8}{\boldmath$\scriptscriptstyle \pm 1.26$}}} &
\makecell{\textbf{40.02}\\[-3.2pt]{\scalebox{0.8}{\boldmath$\scriptscriptstyle \pm 2.92$}}} \\
\bottomrule
\end{tabular}
\end{table}

Ablation studies in Tables~\ref{tab:ablation_mean_percent} and~\ref{tab:backbone_ablation} show clear and consistent contributions from the proposed components, with the largest gains in the clinically relevant mid-slab (resection/union) region. In Table~\ref{tab:ablation_mean_percent}, the proposed Lyapunov-guided trajectory distillation ($T_1$) improves performance over the rectified-flow-only setting and clearly outperforms the MSE-based teacher distillation variant ($T_2$), which indicates that the gain comes from the guidance formulation, not from teacher supervision alone. The full-pipeline ablation also shows a steady improvement as augmentation ($A$), image-space supervision ($I$), trajectory distillation ($T_1$), and resection weighting ($W$) are added, with the largest effect on bone-focused error near the union site (e.g., mid-slab bone MAE decreases from 97.28 to 69.88 HU). Table~\ref{tab:backbone_ablation} further confirms that these gains do not depend on model size: with matched depth and parameter count ($\sim$12M), Res-UNet outperforms SwinUNETR~\cite{hatamizadeh2021swin} both with and without $I$, which justifies our backbone choice.

\section{Conclusion}
This work presents a trajectory-distillation framework for bone remodeling prediction in a low-data, class-imbalanced setting. The model leverages privileged training information from a registration-derived teacher with Lyapunov-guided trajectory supervision. Our proposed approach differs in nature from one-step distillation or mask-based anatomical supervision, with guidance applied over the entire evolution trajectory to promote biologically plausible outputs. Despite the difficulty of predicting long-term graft-host healing from early post-operative geometry alone, the model generated realistic union, nonunion, and partial-union patterns rather than smooth, averaged outputs. These findings suggest that geometry-centered trajectory modeling can capture clinically meaningful remodeling structure even when outcome determinants are only partially observed. Key limitations of our work include the absence of important patient-level covariates (e.g., radiation exposure~\cite{sabiq2024evaluating}), incorporation of exact scan time intervals, and the need for external validation beyond our internal dataset.

\section*{Acknowledgements}
We gratefully acknowledge financial support from the Terry Fox Research Institute and the UBC Friedman Award for Scholars in Health. We also thank the UBC ISTAR Group and the Holland Bone and Joint Research Team at Sunnybrook.

\section*{Disclosure of Interests}
The authors have no competing interests to declare.

\bibliographystyle{unsrt}
\bibliography{Paper-5704}
\end{document}